Pre-Print

# Generative Large Language Models (gLLMs) in Content Analysis:

## A Practical Guide for Communication Research

by

**Daria Kravets-Meinke (Dr.)*** [1]

**Hannah Schmid-Petri (Prof. Dr.)** [1, 2]

**Sonja Niemann** [1]

**Ute Schmid (Prof. Dr.)** [1, 3]

[1] Bavarian Research Institute for Digital Transformation (bidt)

[2] University of Passau

[3] Otto-Friedrich-Universität Bamberg

* Corresponding author

**Funding**

This work was supported by the Bavarian Research Institute for Digital Transformation.

**Abstract**

Generative Large Language Models (gLLMs), such as ChatGPT, are increasingly being used in communication research for content analysis. Studies show that gLLMs can outperform both crowd workers and trained coders, such as research assistants, on various coding tasks relevant to communication science, often at a fraction of the time and cost. Additionally, gLLMs can decode implicit meanings and contextual information, be instructed using natural language, deployed with only basic programming skills, and require little to no annotated data beyond a validation dataset—constituting a paradigm shift in automated content analysis. Despite their potential, the integration of gLLMs into the methodological toolkit of communication research remains underdeveloped. In gLLM-assisted quantitative content analysis, researchers must address at least seven critical challenges that impact result quality: (1) codebook development, (2) prompt engineering, (3) model selection, (4) parameter tuning, (5) iterative refinement, (6) validation of the model's reliability, and optionally, (7) performance enhancement. This paper synthesizes emerging research on gLLM-assisted quantitative content analysis and proposes a comprehensive best-practice guide to navigate these challenges. Our goal is to make gLLM-based content analysis more accessible to a broader range of communication researchers and ensure adherence to established disciplinary quality standards of validity, reliability, reproducibility, and research ethics.

**Keywords**

*content analysis, large language models, automated content analysis, LLM-assisted content analysis, best-practice guide*

# Generative Large Language Models (gLLMs) in Content Analysis: A Practical Guide for Communication Research

Before the rise of computational social science and its methods, human coders were the go-to standard for content analysis in communication research. However, humans bring several limitations: we are slow, costly, biased, inconsistent, and have limited attention spans, limiting the volume of data we are able to analyze manually (Grimmer & Stewart 2013). With the advent of big data, automated methods for content analysis have advanced rapidly in recent years: from simple dictionary-based approaches and word frequencies to supervised and unsupervised machine learning techniques (for overviews, see Boumans & Trilling, 2016; Grimmer & Stewart 2013; Kroon et al., 2024). However, these approaches often struggle with more complex interpretative challenges that require contextual understanding, coding items that go beyond their training data, or the interpretation of implicit meanings such as irony, metaphor, or sarcasm (Kroon et al., 2024). Moreover, while state-of-the-art supervised machine learning models (e.g, DeBERTa) can yield strong performance, they require advanced technical skills and large amounts of high-quality annotated training data (typically ranging between 1,000 and 10,000 texts), which is often hard and costly to acquire (Weber & Reich, 2024; Ollion et al., 2024). As a result, the adoption of supervised machine learning for content analysis in social sciences has remained limited (Chew et al., 2023). The public release of ChatGPT in November 2022—a generative Large Language Model (gLLM) built on transformer architecture and fine-tuned for dialogue using reinforcement learning from human feedback—marked a turning point that could fundamentally transform how data is coded for research purposes (Kuzman et al., 2023; Ziems et al., 2024). Early studies demonstrate that gLLMs can match or even outperform trained and expert human coders across a variety of coding tasks relevant to communication scientists, potentially enhancing or replacing conventional approaches (Gilardi et al., 2023; Heseltine & Clemm von Hohenberg, 2024; Kroon et al., 2024; Ornstein et al., 2024)

Yet, the integration of gLLMs into the methodological toolkit of communication research remains underdeveloped. While a few recent papers offer valuable initial guidance (see Chae & Davidson, 2025; Farjam et al., 2025; Pangakis et al., 2023; Törnberg, 2024a), existing best-practice contributions often assume prior technical knowledge and typically address only selected aspects of the coding pipeline. Against this backdrop, this paper offers an introductory, comprehensive, and literature-based guide specifically tailored to (communication) scholars with no prior experience using gLLMs for quantitative content analysis, but who are interested in applying them in the future. It addresses critical challenges related to (1) codebook development, (2) prompt engineering, (3) selection of the appropriate gLLM, (4) parameter tuning, (5) iterative refinement, (6) validation of the model's reliability, and optionally, (7) enhancement of the model's performance. By synthesizing current research on gLLM-assisted quantitative content analysis, we propose a guiding framework for implementing content analysis with generative AI. Our overall goal is to make content analysis using gLLMs more accessible to communication researchers and to ensure its compliance with established quality criteria for quantitative content analysis, particularly in terms of validity, reliability, reproducibility, and research ethics (Haim et al. 2023; Neuendorf, 2017).

This paper is structured as follows. We begin by summarizing current research on gLLM-assisted content analysis. Here, we focus on "deductive [classification] techniques" (Boumans & Trilling, 2016, p. 9), in which categories of interest are defined *a priori*, typically based on theoretical considerations. Building on insights from this research and our own practical experience with the method, we propose a systematic best-practice approach. Our approach offers brief explanations and suggested solutions to the previously mentioned critical challenges, along with a practical guide tailored to communication researchers, which aims to mitigate the "academic Wild West" (Törnberg, 2024a, p. 1) associated with the current lack of standards for applying gLLMs in quantitative content analysis.

**How LLMs Can Solve Common Challenges in Quantitative Content Analysis**

The potential of gLLMs such as ChatGPT for social science research has recently garnered significant interest both within and beyond the communication science community, with studies demonstrating several advantages of gLLM-assisted content analysis (e.g., Gilardi et al., 2023; Farjam et al., 2025). A meta-analysis of 12 studies confirmed that most found that gLLMs have the potential to enhance the content analysis process (Pavlovic & Poesio, 2025). For example, Törnberg (2024b) showed GPT-4 outperformed experts and crowd-workers in classifying political affiliation of Twitter posts, decoding implicit meanings at lower cost and greater speed (see also Gilardi et al., 2023). Research has also demonstrated that gLLMs perform well across diverse tasks and languages, with only minor drops for longer or non-English texts (Ornstein et al., 2024; Heseltine & Clemm von Hohenberg, 2024). In addition, Kuzman et al. (2023) showed strong performance in under-resourced languages, and Farjam et al. (2025) reported comparable effectiveness when prompt and text languages differed. These findings indicate that gLLM can significantly facilitate multilingual, cross-national, and longitudinal research, and help overcome long-standing barriers in large-scale text analysis (see also Kroon et al., 2024).

Additionally, state-of-the-art gLLMs can be instructed using natural language, deployed with only basic programming skills, and require little to no annotated data beyond a validation dataset—constituting a paradigm shift in automated content analysis (Chae & Davidson, 2025; Kroon et al., 2024). These features substantially lower the technical and resource barriers traditionally associated with automated approaches, enabling researchers to focus more on conceptual, theoretical, and interpretive aspects of their work. As a result, gLLMs have the potential to drastically increase the efficiency and accessibility of automated content analysis in communication research, thereby opening up new possibilities for computational communication science.

At the same time, there are valid concerns that gLLMs operate as black boxes, often lack reproducibility and reliability, raise privacy and ethical concerns, and are prone to bias, some even advising against using gLLMs for substantive content analysis tasks in social science research altogether (Kristensen-McLachlan et al., 2023; Ollien et al., 2023; Reiss, 2023). While recognizing these significant limitations of gLLMs, we argue that many of them can be mitigated through task-specific, rigorous validation and adherence to broader disciplinary quality standards (for similar arguments with regard to computational methods in general, see Song et al., 2020).

**Best-Practice Guide**

While many steps in gLLM-assisted content analysis follow the logic of traditional manual quantitative content analysis (see Krippendorff, 2013; Neuendorf, 2017), certain components diverge from the conventional workflow (for an overview, see Figure 1). Notably, when a study involves coding multiple variables, the full workflow must be repeated—and transparently documented—for each variable individually. To guide researchers through the critical decision-making process in gLLM-assisted content analysis, we summarize our key recommendations for each step in Figure 2.

**Figure 1**

*Workflow for Content Analysis with gLLMs*

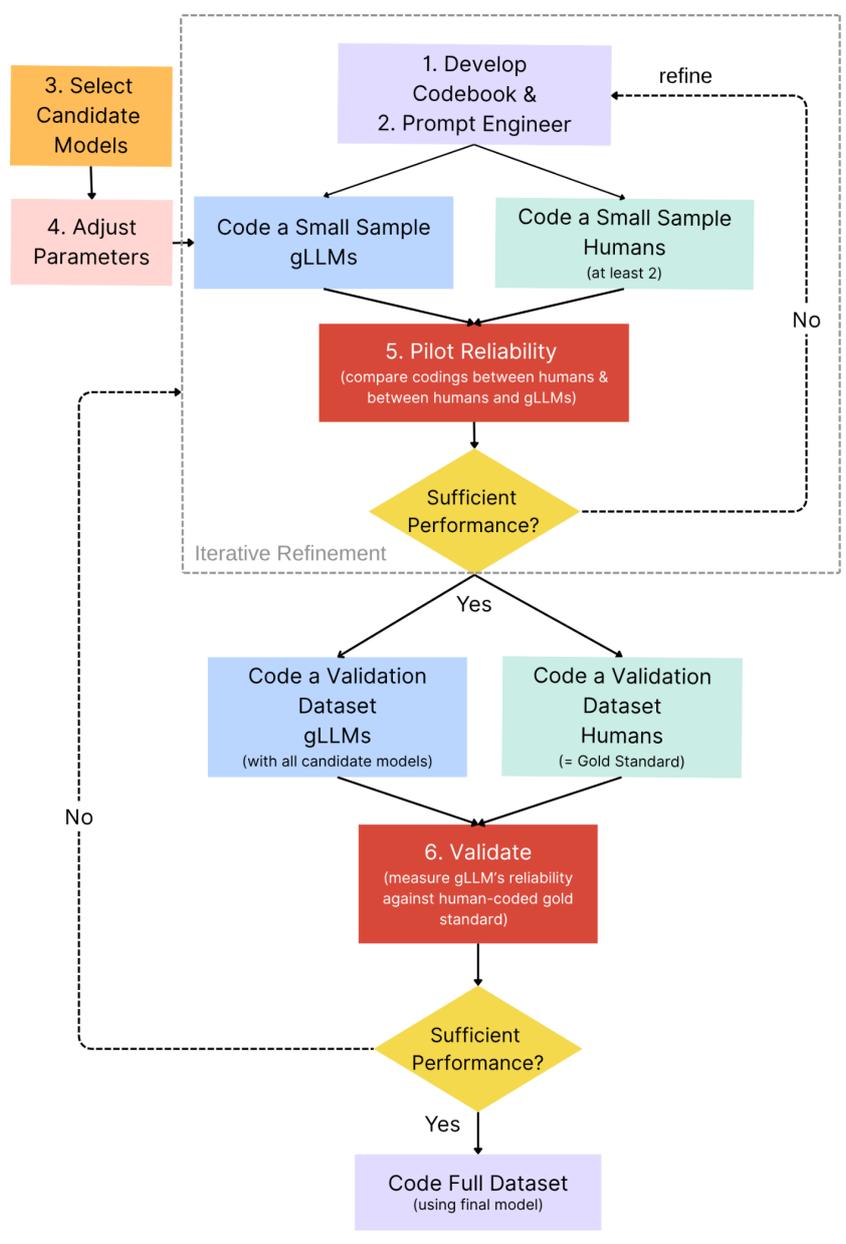

**Figure 2**

*Recommendations for gLLM-assisted Content Analysis*

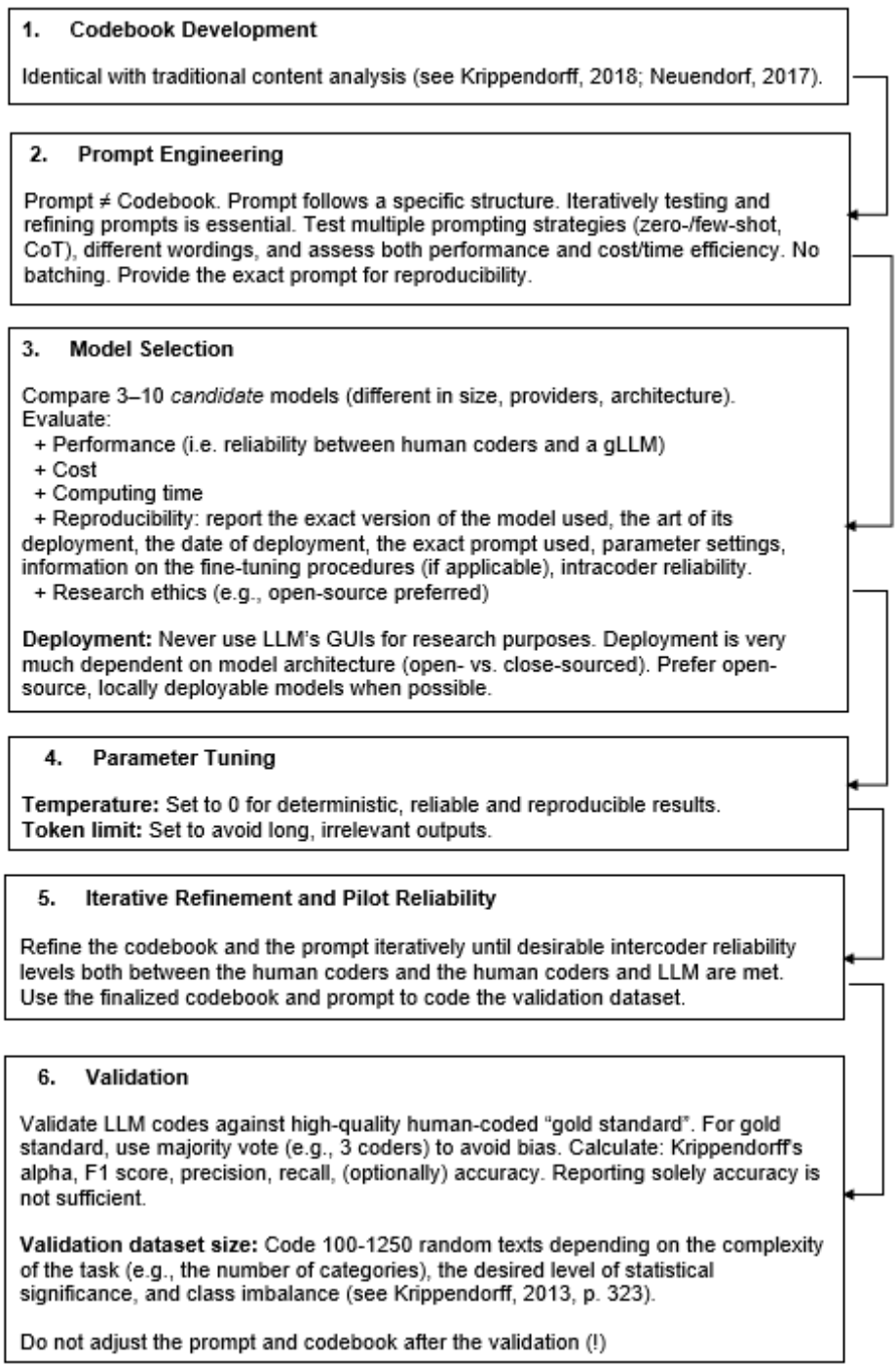

**(1) Codebook Development**

As in traditional manual content analysis, the first step in using gLLMs for data analysis involves developing a comprehensive codebook—a document with clearly defined concepts, categories, coding rules, and examples (Törnberg, 2024a). While creating a codebook for content analysis is a classical task in the social sciences, it is by no means trivial (e.g., Neuendorf, 2017). Codebook creation is an iterative process that requires multiple rounds of testing, refinement, and coder training with human coders to address discrepancies and to ensure that code definitions are well-defined and can be coded consistently (for more detail, see Krippendorff, 2013; Neuendorf, 2017).

**(2) Prompt Engineering**

One distinct task in the process of using gLLMs for quantitative content analysis is prompt engineering. Prompt engineering involves programming gLLMs through written natural language instructions—called prompts—designed to guide how the model codes texts (White et al., 2023). Prompts crucially affect model performance—measured by how accurately the model replicates human-coded "ground truth"—making them a key methodological decision in gLLM-based content analysis (Reiss, 2023; White et al., 2023). Prompt engineering typically proceeds iteratively alongside codebook development. While the concepts, categories, and coding rules must align with those given to human coders, it is not advisable to simply copy the codebook into the prompt, since prompts follow a structure gLLMs process more effectively (e.g., White et al., 2023).

While there is no universally accepted standard how to structure a prompt —and variations exist in the literature, including the ordering of components—a well-structured prompt typically includes several key components (see Figure 3):

(1) *system message* assigning the gLLM's role and situating the task (e.g., "You are a research assistant tasked with analyzing sentiment in texts"; Weber & Reichardt, 2024);

(2) *user message* containing:

    (a) the text to be coded;

    (b) clear instructions defining the coding task, categories, and definitions;

    (c) the desired response format (e.g., "Respond only with a single 1 for yes (mentioned) or 0 for no (not mentioned)"; Weber & Reichardt, 2024; White et al., 2023);

    (d) optionally, coded examples can be added for few-shot learning.

Additionally, research has shown that gLLMs are highly sensitive to wording: small changes—such as "classify" vs. "classification"—can alter performance (Reiss, 2023). Therefore, iterative testing of prompt variations is strongly recommended.

**Figure 3**

*Example of a codebook and a respective zero-shot prompt*

| Codebook | Prompt (zero-shot) |
|---|---|
| You will be provided with tweets about the computer game Borderlands. Please read the entire tweet attentively and determine the tweet's sentiment toward the game between:<br><br>  0 = Negative (The tweet expresses dissatisfaction, criticism, or dislike toward Borderlands)<br>  1 = Positive (The tweet expresses praise, enjoyment, or support for Borderlands)<br>  2 = Neutral (The tweet neither clearly praises nor criticizes Borderlands, or expresses mixed or unrelated opinions)<br><br>Code the representative number. | You are a research assistant tasked with analyzing sentiment in texts. *[role, system message]*<br><br>You will be given individual tweets about the computer game Borderlands. Your task is to determine the tweet's sentiment toward the game using the following coding scheme:<br><br>0 = Negative — The tweet expresses dissatisfaction, criticism, or dislike toward Borderlands.<br>1 = Positive — The tweet expresses praise, enjoyment, or support for Borderlands.<br>2 = Neutral — The tweet neither clearly praises nor criticizes Borderlands, or expresses mixed or unrelated opinions.<br><br>Carefully read each tweet and output only the number that corresponds to the most appropriate sentiment label. *[response format]*<br><br>Tweet: [tweet] *[text to analyze]* *[user message]* |

*Prompting Strategies: Zero-Shot, Few-Shot, and Chain-of-Thought*

Researchers have explored various strategies to optimize prompt performance. Among them, one-shot and few-shot prompting—where the model is provided with one or several coded examples (typically 2–5) directly in the prompt—have received particular attention (White et al., 2023). This approach enables gLLMs to learn from the given examples, often improving performance on specific coding tasks compared to zero-shot approaches (i.e., providing no examples; White et al., 2023).

However, empirical findings on few-shot prompting in content analysis remain mixed. While Weber and Reichardt (2023) and Zhong et al. (2023) report consistent performance improvements across tested models, other studies highlight more variable outcomes. Alizadeh et al. (2025), for instance, observed performance gains in some tasks but declines in others with no clear pattern related to task complexity or model used (for similar findings see Chae & Davidson, 2025; Zendel et al., 2024). Moreover, some researchers warn that few-shot prompts may over-sensitize the model to the examples provided (e.g., Zhong et al., 2023).

Another promising strategy is Chain-of-Thought (CoT) prompting, which encourages the model to provide step-by-step reasoning for its output and has been shown to enhance its problem-solving capabilities (Wei et al., 2022; He et al., 2023). The most basic version of CoT prompting involves appending "Let's think step by step" to the prompt, which has been shown to improve zero-shot performance in content analysis (He et al., 2023). A variation of this approach was noted by Farjam et al. (2025), who found that prompting the model to "quote directly from the text to justify each assigned code" (p. 10) improved reliability scores. CoT can also be combined with one- or few-shot prompting by providing examples together with reasoning. He et al. (2023), for instance, found that this approach outperformed simple variants across three different coding tasks. These gains, however, come at a cost: CoT prompts substantially increase input and output length, raising processing time and financial cost. González-Bustamante (2024), for instance, reports that CoT prompts in their study required over one minute per case, limiting the applicability of such approach for large-scale applications.

Despite these constraints, CoT prompting can be useful during the initial prompt development phase. Asking the model to explain its coding decisions can help researchers identify weaknesses in the prompt and facilitate iterative refinement. Furthermore, some studies suggest that gLLMs can actively assist in prompt refinement by evaluating or revising their own instructions (e.g., Zhong et al., 2023).

In summary, no prompting strategy universally outperforms others (Weber & Reichardt, 2024) and prompt engineering is an iterative and task-specific process. While simple classification tasks may require only minimal instruction, more complex forms of content analysis most likely benefit from richer prompts that include examples or structured reasoning. We recommend that researchers systematically experiment with multiple prompt formats—including zero-shot, one-shot, few-shot, and CoT strategies—while weighing their respective trade-offs in terms of performance (i.e. reliability against human coders), interpretability, cost, and computational time.

*Single vs. Multi-Document Input: Strategies for Batching Texts*

Researchers applying gLLMs to content analysis will inevitably face a key practical decision: Should texts be submitted to the model one at a time, or grouped and processed in batches within a single prompt? While batching multiple texts into one prompt can reduce processing time and lower token usage—thereby cutting costs—it also introduces methodological challenges that may compromise model performance.

The primary concern lies in how gLLMs handle multiple inputs within a single prompt. Due to their reliance on self-attention mechanisms, a model's interpretation of any given input is influenced by its position in the prompt (Lin et al., 2024). This can lead to degraded performance, particularly for items placed in the middle of the batch, where contextual signals are more diffuse (Lin et al., 2024). To mitigate these effects, Lin et al. (2024) propose reordering the same batch multiple times and aggregating outputs via majority vote. In their study, they submitted five permutations of a batch of 32 input texts and found that this approach could match—or even exceed—the performance of single-input prompting, while using fewer tokens overall (see also Zendel et al., 2024). However, the reliability and reproducibility of such techniques remain insufficiently tested across diverse tasks and models.

Given these limitations, we recommend using single-input prompting—processing one text a time. While this approach is more resource-intensive, it minimizes unwanted contextual spillover between entries and ensures more consistent and controlled model behavior.

**(3) Model Selection**

When selecting a gLLM for content analysis, we recommend a two-step approach. In the first step, researchers identify a set of candidate models that are potentially suitable for the coding task (e.g., 3–10). This initial selection can be informed by prior performance on similar tasks and/or practical constraints (see below). In the second step—discussed in more detail in the *Validation* section—all candidate models are benchmarked against human-coded gold standard to assess their actual performance on the coding task at hand.

Although ChatGPT is the most widely recognized gLLM, a growing number of alternative—often open-source—models have demonstrated comparable capabilities (Alizadeh et al., 2025; Weber & Reichardt, 2024; Ziems et al. 2023). Selecting an appropriate model for content analysis should be guided by a critical evaluation that balances disciplinary quality criteria—including validity, reliability, reproducibility, and ethical considerations (Haim et al., 2023; Neuendorf, 2017)—with practical factors such as cost-efficiency and computing time.

In addition, there is a set of hard constraints that determine whether a model is technically viable for the intended coding task. These include (1) language compatibility, (2) context window size, and (3) knowledge cutoff. First, while multilingual gLLMs are increasingly available, not all of them are, and their performance varies substantially across languages. Many models exhibit systematic performance biases, typically favoring English and other high-resource Western languages (Heseltine & Clemm von Hohenberg, 2024). Moreover, some models are known to handle specific languages better than others, making language compatibility a critical consideration in model selection. Second, many gLLMs— particularly older models—limit the length of possible input texts (called *context windows*). For example, LLaMa 2 is limited to approximately 3,000 words, while LLaMA 3 can process up to 96,000 words (Meta, 2024). The recently released LLaMA 4 (April 2025) reportedly supports up to 7,5 million words (Meta, 2025), effectively removing any length constraints for gLLM-assisted content analysis. As such, it is important to consider the length of the documents being analyzed. If the analysis involves longer texts—such as multi-page news articles—it will be necessary to use models with longer context windows (Chae & Davidson, 2025). Third, models also differ in their knowledge cutoffs (i.e. the date of their most recent training data), which can be critical when the coding task relies on contextual information about recent events. A model with a knowledge cutoff in 2022, for example, would not be aware of events, terminology, or developments from 2022 onward.

Finally, we strongly advocate for the use of open-source gLLMs over proprietary models like OpenAI's GPTs. Open models are typically more transparent, cost-effective, and reproducible, and they offer superior standards in terms of data privacy and long-term accessibility (see Alizadeh et al., 2025; Ollion et al., 2024; Spirling, 2023). In the following, we summarize the core quality criteria researchers should consider when selecting an gLLM for content analysis in communication research.

*Performance*

A model's performance against high-quality—reliable and valid—human-coded "ground truth" is a seminal quality criterion for the validation of automated content analysis (Grimmer & Stewart, 2013; Haim et al., 2023). It is also undoubtedly a critical factor in its selection, as capabilities can vary significantly across models and tasks. In this regard, comparative evaluations consistently show that there is no universally superior model. For instance, Pilny et al. (2024) compared the performance of five popular gLLMs *(gpt-4.0-turbo, gpt-3.5-turbo, Claude 2, llama7b-v2-chat*, and *llama13b-v2-chat*) across multiple content analysis tasks and found that performance was task-dependent. Larger, more complex or newer models did not necessarily outperform smaller, simpler or earlier ones (see also Chew et al., 2023; González-Bustamante, 2024). Importantly, open-source models can also perform on par with proprietary ones. Alizadeh et al. (2025), for instance, demonstrated that open-source models such as LlaMa and Flan achieved comparable performance levels to GPT-3.5 on specific annotation tasks (for similar findings, see Chae & Davidson, 2025; Kristensen-McLachlan et al., 2023).

In short, there is no one-size-fits-all solution, and model selection should be tailored to the specific coding task. While model performance remains essential and serves as a knockout criterion, ideally, it should not be the sole consideration guiding model selection. Instead, if multiple models reach sufficient levels of performance, we advocate for a balanced evaluation that considers cost-efficiency, reproducibility, and ethical considerations alongside performance (for a similar argument, see Chae & Davidson, 2025). For our recommendations regarding the appropriate size of the validation dataset, suitable validation metrics, and the construction of a high-quality "ground truth," see the *Validation* section.

*Reproducibility*

Reproducibility is a fundamental criterion of scientific research. It refers to the ability of independent researchers to replicate a study's procedures and obtain comparable results using the same data and methods (Neuendorf, 2017). Reproducibility is thus determined by full disclosure of information on methodology and procedures. In the context of gLLM-assisted content analysis, this includes: the exact model version, deployment method and date, exact prompt used, parameter settings, fine-tuning procedures (if applicable), size of the validation dataset, and its relation to the overall sample (including how it was selected), sampling strategy, codebook, number of human coders and their intercoder reliability scores, as well as gLLM's validation performance (see *Validation*) and intracoder reliability (see *Parameter tuning*).

The reproducibility of proprietary models is typically low (Alizadeh et al., 2025; Ollion et al., 2024; Spirling, 2023). Such models are often updated without user notification, and research has shown that seemingly the same model yielded different results only weeks apart (Ollion et al., 2024). Moreover, proprietary providers like OpenAI frequently deprecate older models without archiving them, thereby undermining long-term reproducibility (Liesenfeld et al., 2023; OpenAI, 2025c). In this regard, open-source models offer a clear advantage: they can be downloaded, stored, and archived, enabling researchers to preserve the exact model used and facilitate replication over time. However, all gLLMs—open or closed—share a fundamental limitation: limited transparency and explainability. These models are typically trained on undisclosed datasets (for a detailed analysis of how open open-sourced models are, see Liesenfeld et al., 2023) and operate as black boxes, making it difficult to trace specific outputs to particular inputs or training data. Notably, asking a model to explain its coding decision (e.g., via CoT prompting) does not guarantee insight into its actual reasoning processes. Rather, such explanations may be generated to appear plausible to the user, without offering genuine insight into the model's internal decision-making logic (see Zendel et al., 2024).

*Cost-efficiency*

Large-scale content analysis using gLLMs can be costly—particularly when relying on proprietary models such as OpenAI's GPTs, which typically operate on a pay-per-token basis. *Tokens* refer to chunks of text (e.g., words or word parts) used to calculate input and output length; longer prompts and responses therefore result in higher usage costs. Tokenization schemes vary across providers, but in OpenAI's models, one token roughly corresponds to 0.75 words of English text (OpenAI, 2025a). While the price per token is typically low—for instance, GPT-4o charged $2.50 per one million input tokens (approximately 750,000 words) at the time of writing (OpenAI, 2025b)—total costs can quickly accumulate depending on text length, number, and the prompt. In this regard, open-source gLLMs present a more accessible alternative, as these are free to use, with expenses limited to computational resources. However, larger models often require the use of commercial cloud infrastructure, which can incur substantial expenses. As such, both proprietary and open-source solutions come with distinct cost structures that should be carefully compared during model selection.

*Research Ethics*

Model selection in gLLM-assisted content analysis must also meet the standards of research ethics, particularly regarding openness, data privacy, and accountability. Open research practices are important drivers of scholarly research (e.g., Liesenfeld et al., 2023).

Proprietary models are closed systems: their training data remain undisclosed, their architectures are opaque, and their updates are undocumented, limiting independent inspection and verification (Spirling, 2023). Moreover, user data entered into proprietary models may be retained or used for further training, raising concerns around informed consent and data protection—especially when dealing with sensitive or regulated content (e.g., under the GDPR; Ollion et al., 2024; Spirling, 2023). In contrast, open-source gLLMs typically offer greater transparency and user control: researchers can inspect or modify the source code, audit model behavior, and ensure that data are processed locally (Liesenfeld et al., 2023). Furthermore, open-source models can potentially foster more diverse and decentralized development communities, which may improve the overall fairness and robustness of these models compared to proprietary gLLMs (Alizadeh et al., 2023).

In addition, researchers should also consider the provenance and governance context of the models they use. Models developed or funded under authoritarian regimes, such as *DeepSeek* or *Qwen* from China, may raise concerns about the values embedded in model development and deployment. As AI infrastructure becomes geopolitically consequential (see Jungherr, 2024), academic researchers have a responsibility to prioritize models aligned with democratic principles, transparency, and civic accountability.

Finally, the environmental footprint of gLLM use warrants ethical scrutiny. Although recent studies suggest that gLLMs may be less resource-intensive than manual coding on the same task (Ren et al., 2023), large-scale deployments remain computationally demanding (Palicki et al., 2025). Therefore, while analyzing full datasets may seem attractive—especially given the speed and scalability of gLLMs—researchers should question whether full dataset analysis is necessary or whether sampling-based approaches may suffice (see boyd & Crawford, 2012). Additionally, research has demonstrated that smaller models are more sustainable than larger and more complex models (Palicki et al., 2025). We therefore recommend a critical assessment of model size relative to the coding task complexity and encourage the selection of smaller models wherever feasible.

*Computing Time*

Computing time is another important consideration when selecting gLLMs for content analysis. While less important for smaller datasets, computing time can become an issue for larger datasets. For example, González-Bustamante (2024) compared the processing time of various gLLMs and human crowdworkers in coding the toxicity of 1,000 Spanish tweets (binary classification). He found that GPT-4 completed the task in approximately 14 minutes—processing each tweet in 0.852 seconds—making it five times faster than the average of ten tested open-source models, which were, in turn, nearly six times faster than human coders. However, when extrapolated to González-Bustamante's (2024) full dataset of 3.5 million tweets, even GPT-4 would require an estimated 828 hours (or 35 days) to complete the task, with open-source models, being five times slower, needing significantly longer. We therefore recommend that researchers record and compare computing time necessary to process the validation dataset (or as an average per item) between the models. While computing time is unlikely to be a primary constraint in most applications—and will likely continue to decrease as gLLM technology advances—it may serve as a relevant secondary criterion when models perform similarly on key dimensions such as accuracy, cost, or reproducibility.

*Deployment of the Model*

Deployment strategies for gLLMs—that is, the way researchers access and execute the models—are closely tied to core research quality standards and are important factors to consider when selecting a model. Researchers typically have three options for deploying gLLMs for content analysis, each with distinct implications: graphical user interfaces (GUIs), application programming interfaces (APIs), and local deployment.

**GUI-Based Interfaces.** For many users, gLLMs are primarily encountered through web chat interfaces such as ChatGPT. But can such interfaces be reliably used for systematic data analysis? Heseltine and Clemm von Hohenberg (2024), for instance, conducted their entire data analysis using GPT-4 on a standard $20 ChatGPT subscription, available at the time, and found their results broadly aligned with expert-coded "ground truth." Indeed, it is technically possible to analyze entire datasets—albeit limited in size—by uploading them directly into gLLM chats. However, this approach is highly problematic and we strongly advise against using GUI-based interfaces for systematic content analysis in research contexts:

1. **Data privacy concerns**: Companies such as OpenAI explicitly state that user data submitted through chat interfaces may be stored and used for further model training (OpenAI, 2025d). This poses significant risks when handling sensitive or regulated data, such as copyrighted materials or personally identifiable information (Ollion et al., 2024).
2. **Lack of parameter control**: GUI interfaces typically do not allow researchers to adjust key model settings (e.g., temperature), which are essential for ensuring consistent and reproducible outputs.
3. **Limited scalability and transparency**: gLLM chats are designed for manual, one-off interactions, not for analyzing large datasets. Manually copy-pasting texts into chats is not a practical way to conduct automated content analysis. While some platforms support uploads of structured datasets (e.g., CSV,

Excel), their size is limited and opaque processing logic raises concerns about transparency and replicability.

In sum, while GUI-based interfaces may be helpful during early-stage experimentation or prompt development, they are not suitable for systematic content analysis in communication research.

**APIs.** Instead, APIs offer a practical and widely used solution for employing gLLMs in content analysis, enabling structured, automated interactions with both proprietary and (externally) hosted open-source models. The typical procedure involves researchers sending prompts programmatically via API requests—typically via R or Python—to which the model responds with structured outputs.

A standard API call (see Figure 4) includes the model endpoint, a prompt, optional parameters, an authentication token (API key), and a response object containing the model's output. While the exact structure of the call may vary by model and provider, the general logic remains the same: researchers iterate through a dataset, sending one prompt per text and capturing the response in a structured format. We strongly encourage researchers to share their data analysis code including the API call to ensure transparency, reproducibility, and comparability across studies.

**Local Deployment.** Deploying a gLLM locally is often considered the gold standard in terms of reproducibility, data privacy, and long-term reliability (e.g., Alizadeh et al., 2025; Weber & Rechardt, 2024). Running a model on one's own infrastructure offers full control over the computing environment, avoids dependencies on third-party APIs, and mitigates risks associated with platform changes, usage restrictions, or paywalls that might hinder replication efforts in the future (Ollion et al., 2024). However, this approach also comes with specific technical and hardware requirements that may not be feasible for all researchers.

**Figure 4**

*Example Python API Call Using OpenAI's openai Library*

```python
import openai

openai.api_key = 'your-api-key-here'

response = openai.ChatCompletion.create(
    model="gpt-4",
    messages=[
        {"role": "system", "content": "You are a research assistant."},
        {"role": "user", "content": "Classify this text as positive or negative: 'I love this product'"}
    ],
    temperature=0,
    max_tokens=300
)

print(response['choices'][0]['message']['content'])
```

*Note.* Instead of requiring a manual endpoint URL, the openai library handles communication with OpenAI models internally.

The primary constraint is computational capacity. State-of-the-art models consist of billions of parameters, and the larger the model, the greater the memory and processing demands. As a general rule of thumb, local deployment requires approximately 2 GB of GPU memory (VRAM) per billion parameters. For instance, a 7-billion-parameter (7B) model requires at least 14 GB of VRAM (Alizadeh et al., 2025), which exceeds the capacity of most consumer-grade computers. To reduce hardware demands, researchers often use quantization, which compresses model weights into lower-precision formats (e.g., 4-bit or 5-bit), significantly reducing memory usage with minimal performance loss (Dettmers et al., 2023). For instance, a quantized LLaMA 7B model can be run on modern high-end laptops or desktops.

Tools such as *llama.cpp*, *GPTQ*, or *Ollama* simplify local deployment. Ollama, in particular, provides a user-friendly interface for running gLLMs locally via the command line or Python, enabling scalable content analysis with full data control (Gruber & Weber, 2024).

Still, local deployment requires at least intermediate technical expertise, including familiarity with Python, command-line tools, and model configuration. Additionally, not all models can be deployed locally: while open-source models (e.g., LLaMA, Mistral, Gemma) support local inference, proprietary models like GPTs or Claude are only accessible via APIs or GUIs.

To sum up, by the end of the first model selection step, researchers should identify approximately 3 to 10 *candidate* models deemed suitable for their content analysis task (e.g., based on their past performance on a similar task or hard contrains)—testing between different parameter sizes (e.g., LLaMA 7B vs. 70B), providers (e.g., OpenAI vs. Google), and openness (open- vs. closed-source).[1] In the subsequent *Validation* step, the performance of these candidate models is evaluated on the specific content analysis task at hand.

**(4) Parameter Tuning**

We recommend that researchers configure several key parameters when using gLLMs for content analysis (see Figure 3 for implementation details):

1. **Temperature:** among all model parameters, temperature is arguably the most critical. It controls the randomness in the model's output: lower values (e.g., 0 to 0.2) make the model more deterministic by encouraging it to select the most probable next token. This is particularly important for content analysis, where reliability and reproducibility are essential. Research has demonstrated that low-temperature settings are generally more suitable for content analysis (Alizadeh et al., 2025; Gilardi et al., 2023; Reiss, 2023; Ziems et al., 2023). For instance, González-Bustamante (2024) reports near-perfect intracoder

---

[1] A practical way to compare the performance of multiple (open-source) gLLMs without the need to set them up locally—at least for the validation step—is to use commercial production-ready cloud infrastructure (e.g., *DeepInfra*, *Replicate*, *OpenRouter*) that allows access to multiple gLLMs through one API.

reliability between multiple gLMM iterations when the temperature was set to zero (see also Reiss, 2023).[2]

2. **Token limit**: this parameter defines the maximum number of tokens a model may return in a response. While token limit setting does not influence the nature or quality of the model's output, it solely cuts the response down to $n$ tokens if the gLLM intended to generate a longer response. Because many API-based services charge per token—and token generation can be slow—setting a reasonable limit improves both speed and efficiency. Importantly, it also serves as a safeguard: when gLLMs drift from the assigned coding task or produce irrelevant content (as occasionally observed; see Pilny et al., 2024), shorter responses limit the scope and potential impact of such off-target outputs.

3. **Response format:** some gLLMs allow users to specify a structured response format. For example, OpenAI's GPT models offer a structured outputs feature, which enables users to define a fixed JSON schema for the model's responses (OpenAI, 2025e). When available, this feature helps enforce consistency in output structure, which is useful for automated parsing and downstream analysis.

**(5) Iterative Refinement and Pilot Reliability**

Next, a small random sample (e.g., 50 texts) is drawn from the dataset and coded by both the selected gLLMs and at least two trained human coders (see *Validation*; Song et al., 2020; Neuendorf, 2017). The results are then compared, and discrepancies—whether among human coders (e.g., during coder training) or between coders and the gLLMs (e.g., by prompting the model to explain its decisions)—are identified and resolved. Based on these comparisons, both the codebook and the prompt are revised—clarifying category definitions, adjusting instruction phrasing, or updating examples. This iterative process is repeated until the desired performance thresholds are met, and further refinements yield no substantial improvements. The finalized codebook and prompt are then used to code the validation dataset.

---

[2] Nevertheless, González-Bustamante (2024) found that among the models tested, one model—GPT-4— failed to reach acceptable levels of intracoder reliability on their task, even with the temperature parameter set to zero. We therefore encourage researchers to test and report intracoder reliability in gLLM-assisted content analysis, as it represents a critical dimension of overall reliability.

**(6) Validation**

Validity is a fundamental quality criterion in content analysis. It determines whether the results accurately reflect an underlying empirical truth (e.g., Krippendorff, 2013; Song et al., 2020). In communication science, this truth is typically understood as what humans derive from the content being analyzed (Haim et al., 2023). Accordingly, gLLM-generated codings must be rigorously validated against high-quality human codes (Grimmer & Stewart, 2013; Song et al., 2020; Pangakis et al., 2023). Therefore, to select the final model, researchers must evaluate the performance of all candidate models against a validation dataset, which—together with other established quality criteria in communication research—serves as the basis for final model selection.

*Size of a Validation Dataset*

As in classical quantitative content analysis, there is no universally accepted standard for the size of a validation dataset in gLLM-assisted content analysis. A common rule of thumb suggests allocating between 1% and 20% of the full dataset to validation. In practice, however, this threshold is rarely met, especially in large-scale automated analyses (Song et al., 2020). Instead, researchers recommend using 100 to 1,250 randomly sampled texts, with the exact size depending on task complexity, class imbalance, and the desired confidence level (Pangakis et al., 2023; Neuendorf, 2017).

For example, Alizadeh et al. (2025) recommend that the validation dataset must include at least 100 items and a minimum of two instances of the minority class. If these conditions are not met, the sample should be incrementally expanded until they are. In their study—however, based on relatively small datasets (e.g., up to 2,480 news articles)—they allocated 15% of the dataset to validation and increased this share in 5% steps until the criteria mentioned above were met. These suggestions follow the rough guidelines provided in classical content analysis methods research. Neuendorf (2017, p. 187), for instance, suggests allocating at least 10% of the full dataset to validation, but notes that it rarely needs to exceed 300 cases. For detailed guidelines, including a reference table for calculating the required size of the validation dataset based on significance levels and variable complexity, see Krippendorff (2013, p. 323).

In constructing a validation dataset, researchers must decide between two main sampling strategies: (1) *probabilistic sampling*—that is, randomly selecting items from the full dataset—and (2) *non-probabilistic* (or *rich range*) *sampling*, which involves deliberately including examples of all possible values of the target variable (Neuendorf, 2017). While rich range sampling is not widely recommended for validation purposes in classical quantitative content analysis (Neuendorf, 2017), a hybrid strategy may be practical in gLLM-assisted analysis: for instance, using a rich range sample for coder training and prompt refinement, and a random sample for model validation.

### *Creating a Gold Standard*

Several approaches have been proposed in the literature for comparing gLLM-generated codes to those produced by human coders. One strategy treats the gLLM as an additional coder within a pool of human coders, calculating pairwise intercoder reliability between the gLLM and each human coder individually (e.g., Zendel et al., 2024). More commonly, however, gLLMs are evaluated against a single, consolidated set of human codes treated as the gold standard or "ground truth." To construct this gold standard, researchers typically (1) retain only cases that all coders agreed upon (e.g., Alizadeh et al., 2025); (2) ask coders to resolve their discrepancies (e.g., Chae & Davidson, 2025); or (3) determine the final code by majority vote among an uneven number of coders (e.g., Farjam et al., 2025).

However, not all of these approaches are equally advisable from a methodological standpoint. Krippendorff (2013), for instance, strongly cautions against using consensus coding (i.e. coders discussing their discrepancies) for reliability assessment because it violates one of the ground requirements of reliable content analysis—independent judgment by the coders. Moreover, consensus-based coding can introduce the biases of the group dynamics between the coders, as coders typically "yield to each other in tit-for-tat exchanges, with prestigious group members dominating the outcome" (p. 273). Similarly, relying solely on cases that all coders agreed upon can distort validation outcomes by excluding more complex or ambiguous cases, thus artificially inflating the model's performance metrics. This selective sampling also violates the principles of probability sampling in the construction of validation datasets, which is central to content analysis (e.g., Neuendorf, 2017; Krippendorff, 2013).

Importantly, human coding does not inherently constitute an infallible gold standard. In fact, computer science research has long questioned the reliability of human-coded benchmarks (e.g., Northcutt et al., 2021, in communication science, see Song et al., 2020). One concern is the so-called "expertise paradox": gLLMs, trained on vast and diverse data, may occasionally surpass human coders in their level of expertise. For example, van Hoof et al. (2025) found that GPT-3.5 correctly identified political search queries—such as one about a local politician's relationship status—which all human coders failed to recognize. Another concern is that there is not always a single, definitive way to code a given text (e.g., Baden et al., 2023). Content ambiguities can lead to disagreements between human coders—and between coders and gLLMs—which do not necessarily indicate measurement error but reflect the true nature of the coded text (Baden et al., 2023).

Given these concerns, we recommend constructing the gold standard by employing an uneven number of coders (e.g., three), coding independently, with final codes determined by majority vote. While this approach is not immune to error—agreement does not guarantee correctness—it preserves coder independence and aligns with probabilistic sampling principles.

*Validation Metrics*

Importantly, only variables that achieve sufficient levels of intercoder reliability should be used to validate gLLM-generated outputs. What constitutes *sufficient* agreement remains a subject of scholarly debate (Neuendorf, 2017; Krippendorff, 2013). As a rule of thumb, Krippendorff's *alpha*[3] values above .80 are considered generally sufficient. However, for more complex or abstract constructs, lower thresholds—typically between .667 and .80—may be acceptable for drawing tentative conclusions (Krippendorff, 2013, p. 325; Neuendorf, 2017).

Once a human-coded gold standard has been established, the performance of each candidate gLLM should be evaluated against it using established validation metrics. Among the most widely applied metrics are precision, recall, and the F1 score,[4] as well as

---

[3] While not the only available coefficient for measuring agreement in content analysis, *Krippendorff's alpha* is generally considered the standard due to its flexibility—in particular, with regard to the number of coders, scales of measurement (nominal, ordinal, interval, and ratio), missing data, sample sizes—and theoretical rigor. Alternative coefficients include, but not limited to, Cohen's kappa and Fleiss' kappa.

[4] Alternative metrics include MCC, ROC and AUC (e.g., He & Garcia, 2009).

Krippendorff's alpha and accuracy (Song et al., 2020; Panagakis et al., 2023; Weber & Reichardt, 2024). Precision measures the proportion of predicted positives that are actually correct, while recall (or sensitivity) captures the proportion of actual positives correctly identified by the model. Depending on the research objective, one may be prioritized—for instance, recall for identifying all relevant cases, or precision for minimizing false positives. To balance both, the F1 score—the harmonic mean of precision and recall—is commonly used, especially when working with datasets, where one class is far more frequent than others. For multi-categorical variables, aggregated F1 scores are commonly used: (1) *macro* F1 averages the F1 scores for each category equally, regardless of category size; (2) *weighted* F1 adjusts for category imbalance by weighting each category according to its frequency in the dataset.

In contrast to Krippendorff's alpha values, there is no recommended threshold for acceptable precision, recall, or F1 score. Instead, researchers stress that the appropriate benchmark depends on the complexity of the coding task, class distribution, and overall research objectives (Grimmer et al., 2022). Consequently, we recommend assessing multiple validation metrics—including precision, recall, F1 (macro and weighted, where applicable), and, optionally, accuracy and Krippendorff's alpha—to provide a comprehensive assessment of model performance. Finally, accuracy alone is not a reliable performance metric, especially when working with imbalanced datasets, as high accuracy can obscure poor performance on minority classes by merely capturing the model's ability to predict the dominant class (He & Garcia, 2009; Krippendorff, 2013).

**(7) Performance Enhancement**

In more complex cases—where gLLM-generated codes fail to meet sufficient validation thresholds, even after multiple rounds of prompt optimization and despite acceptable intercoder reliability among human coders—two promising strategies for enhancing model performance are: (1) hybrid coding and (2) fine-tuning.

**(1) Hybrid coding** refers to the content analysis process where gLLMs handle high-confidence classifications, while ambiguous or complex instances are deferred either to human coders (Heseltine & Clemm von Hohenberg, 2024) or to a secondary "judge" gLLM (Farjam et al., 2025, p. 4). In practice, one way to implement hybrid coding is to let the gLLM assign codes to texts along with confidence scores, so that high-confidence codes can be accepted automatically, whereas low-confidence cases require manual review. Alternatively, the model—or even multiple models—may code the data multiple times (e.g., using a higher temperature setting), and in cases where the codes diverge across iterations, a human or judge LLM interferes for the final decision. Research has shown that pooling codes from multiple iterations can improve the overall performance of gLLMs in content analysis (Farjam et al., 2025; Reiss, 2023). While such human-in-the-loop approaches require more manual coding, they still substantially reduce the overall effort required for content analysis.

**(2) Fine-Tuning** refers to retraining a gLLM on task-specific annotated data to improve domain performance (Alizadeh et al., 2025). Typically, researchers provide 50–250 coded examples to help the model learn task-specific patterns. Another form of fine-tuning is instruction tuning, which uses large instruction–input–output datasets to strengthen the model's ability to follow natural language instructions and has been shown to enhance performance in content analysis (Chae & Davidson, 2025).

Despite its benefits, fine-tuning large models can be computationally demanding, often requiring large-scale GPU clusters (Alizadeh et al., 2025). Recent advances, however, such as quantization and parameter-efficient methods like LoRA (Low-Rank Adaptation), allow fine-tuning even on single-GPU machines (Hu et al., 2023). Fine-tuning is mainly used with open-source models, though some proprietary providers, such as OpenAI, also offer limited options (OpenAI, 2025b). Yet in many cases fine-tuning may not be unnecessary, as numerous pre-fine-tuned open-source models are already available on platforms likeHugging Face,[5] reducing the need for custom training—especially for common or generalizable coding tasks.

Finally, the evidence on the effectiveness of fine-tuning for content analysis is mixed. While some studies highlight it as a key method for boosting model performance (e.g., Alizadeh et al., 2025) and demonstrate that fine-tuning smaller models can match the performance of larger models at lower cost (Chae & Davidson, 2025), fine-tuning does not always lead to better performance (Ziems et al., 2024) and advances in base model capabilities suggest that the necessity of fine-tuning may decline over time. Given the trade-offs—including computational cost, potential for bias, and required technical expertise—we recommend that researchers first compare the performance of multiple candidate models. If one already meets the validation criteria, we advise foregoing fine-tuning.

## Discussion

The technological advancement of gLLMs has opened up new promising avenues for content analysis in communication science. Research has demonstrated how these models can be effectively applied to a wide range of content analysis tasks in social sciences (Gilardi et al., 2023; Ziems et al., 2023). However, as with other text-as-data methods, gLLM-assisted quantitative content analysis presents at least seven major challenges that researchers using the method must address to ensure result quality: (1) creating a codebook, (2) prompt engineering, (3) selecting the appropriate gLLM, (4) tuning parameters, (5) iteratively refining the codebook and prompt, (6) validating the model's reliability, and (7) enhancing the model's performance (e.g., through hybrid coding). In this paper, we synthesize the growing body of literature on gLLMs in quantitative content analysis addressing these challenges and propose a comprehensive methodological best-practice approach for reliable, reproducible, ethical, and valid gLLM-assisted content analysis of large amounts of textual data in communication science research (our key recommendations are summarized in Figure 2). Our overall goal is to make content analysis using gLLMs more accessible and applicable for a broader range of communication researchers.

---

[5] https://huggingface.co/

That said, gLLM-assisted content analysis is certainly not a silver bullet solution to all content analytic problems. In this, we agree with Grimmer & Stewart (2013): there is no universally best method for automated content analysis. gLLM-assisted content analysis is well suited to contexts when annotated data is scarce, researchers lack computational expertise, or few entities need analysis. However, this does not preclude its use in large-scale applications—especially when researchers have access to substantial computational resources, funding, or technical expertise that allow for the deployment of open-source models or pay-per-token proprietary APIs. Despite current limitations—such as a lack of transparency, dependence on proprietary infrastructure, and uncertainty about training data provenance—gLLMs remain a promising tool for communication research. We argue that many of these challenges may diminish or evolve in the near future and, in the meantime, can be mitigated through rigorous, task-specific validation and a consistent commitment to established disciplinary standards of validity, reliability, reproducibility, and research ethics. Looking ahead, we join others in calling for the development and adoption of open-source gLLMs, driven by academic communities rather than commercial interests, and dedicated institutional infrastructure, support and training to ensure their accessibility (e.g., Ollion et al., 2024). As Spirling (2023) suggests, what the field needs is not just better tools, but coordinated, large-scale initiatives—akin to what CERN is for physics—to develop and maintain gLLMs that serve the specific needs and values of the social sciences.

**Disclosure statement**

The authors report there are no competing interests to declare.